\documentclass[10pt,twocolumn,letterpaper]{article}

\usepackage{iccv}
\usepackage{times}
\usepackage{epsfig}
\usepackage{graphicx}
\usepackage{amsmath}
\usepackage{amssymb}
\usepackage{booktabs}
\usepackage{nicefrac}
\usepackage{mathtools}
\usepackage{multirow}
\usepackage{algorithm}
\usepackage{algorithmic}
\newcommand\myeq{\stackrel{\mathclap{\normalfont\mbox{\scriptsize{STE}}}}{=}}

\usepackage[breaklinks=true,letterpaper=true,colorlinks,bookmarks=false]{hyperref}

\iccvfinalcopy 

\def\iccvPaperID{4601} 

\begin{document}

\title{BitSplit-Net: Multi-bit Deep Neural Network with Bitwise Activation Function}

\author{Hyungjun Kim, Yulhwa Kim, Sungju Ryu, and Jae-Joon Kim\\
Department of Creative IT Engineering, Pohang University of Science and Technology\\
{\tt\small \{hyungjun.kim, yulhwa.kim, sungju.ryu, jaejoon\}@postech.ac.kr}
}

\maketitle

\begin{abstract}
   Significant computational cost and memory requirements for deep neural networks (DNNs) make it difficult to utilize DNNs in resource-constrained environments. Binary neural network (BNN), which uses binary weights and binary activations, has been gaining interests for its hardware-friendly characteristics and minimal resource requirement. However, BNN usually suffers from accuracy degradation. In this paper, we introduce “BitSplit-Net”, a neural network which maintains the hardware-friendly characteristics of BNN while improving accuracy by using multi-bit precision. In BitSplit-Net, each bit of multi-bit activations propagates independently throughout the network before being merged at the end of the network. Thus, each bit path of the BitSplit-Net resembles BNN and hardware friendly features of BNN, such as bitwise binary activation function, are preserved in our scheme. We demonstrate that the BitSplit version of LeNet-5, VGG-9, AlexNet, and ResNet-18 can be trained to have similar classification accuracy at a lower computational cost compared to conventional multi-bit networks with low bit precision ($\leq$ 4-bit). 
   We further evaluate BitSplit-Net on GPU with custom CUDA kernel, showing that BitSplit-Net can achieve better hardware performance in comparison to conventional multi-bit networks.
\end{abstract}

\section{Introduction}
Low-precision Deep Neural Networks (DNNs) are attracting interest as an approach to implement state-of-the-art networks in resource-limited environments such as mobile applications \cite{BNN, gupta, mobile, QNN, WRPN, XNOR}. To reduce the memory requirements and the computing costs of DNNs, various schemes have been proposed \cite{Deepcomp,TWN,ABC,WRPN}. Among them, quantization of the weights and/or activations is one of the most popular approaches. For example, in Binary Neural Network (BNN), both weights and activations are quantized to 1-bit so that memory and computing costs are reduced to a minimum \cite{BNN}. However, BNN tends to suffer a substantial loss in inference accuracy for large neural networks. To reduce the accuracy loss, multi-bit neural networks with 2 to 4-bit precision for weights and activations have been proposed \cite{ABC,WRPN,DoReFa}. While the increased bit precision of weights and/or activations improves classification accuracy, various hardware-friendly features of BNN cannot be applied to multi-bit neural networks because most of the hardware-friendly features stem from bitwise operations.

In this paper, we present a method of maintaining the hardware-friendly features of BNN while achieving a higher accuracy than BNN allows, by using multi-bit precision. In essence, the proposed BitSplit scheme splits a $k$-bit activation into $k$ binary activations. Two main characteristics of BitSplit-Net are as follows: \textbf{(1) each bit of multi-bit activations propagates independently throughout the network, (2) BitSplit-Net uses binary activation functions only;} thus, similar to BNN, the hardware-friendly thresholding function can be used in the proposed scheme. Both characteristics are hardware-friendly in different points of view, thus BitSplit-Net is more hardware-efficient than conventional multi-bit networks.

For verification, we evaluate the BitSplit scheme in both software and hardware levels. We first applied the BitSplit scheme to various networks on different image classification datasets. We demonstrate that the classification accuracy of BitSplit-Net increases as the bit precision of activation increases. This indicates the proposed scheme has good bit-scalability. We further show that the BitSplit version of a network can be trained to have the same level of accuracy as that of its original counterpart. For example, BitSplit versions of AlexNet \cite{AlexNet} and ResNet-18 \cite{resnet} showed little accuracy difference ($<\pm$0.5\%) compared to state-of-the-art quantized neural networks \cite{HWGQ,BNN,QNN,ABC,XNOR,DoReFa}. We also evaluate the hardware benefit of BitSplit scheme on GPU platforms. For the state-of-the-art GPU environment, we implemented custom CUDA kernel for BitSplit scheme. We evaluate the speed-up of BitSplit kernel on matrix-vector multiplication and forward propagation of MLP on MNIST dataset. As a result, BitSplit kernel achieved 4.26$\sim$12.5x speed-up for matrix-vector multiplication and 2.65$\sim$5.89x speed-up for inference on MNIST benchmark. 

\section{Motivation}

\subsection{Does the second-most significant bit alone have enough importance?}

\begin{figure}[ht]
    \centering
    \includegraphics[width=\linewidth]{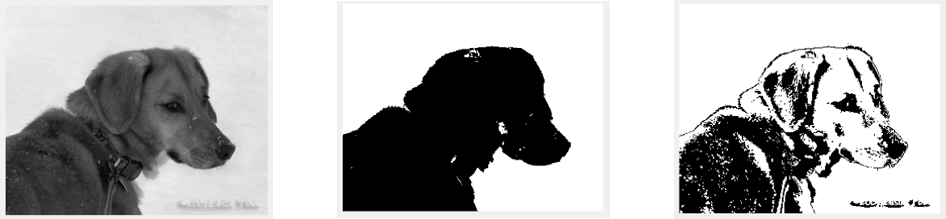}
    \caption{An image from ImageNet dataset in (a) 8-bit, (b) 1-bit (MSB), and (c) 1-bit (2nd MSB) format}
    \label{dog}
\end{figure}

Recently proposed quantized neural networks use limited bit precision for weights and/or activations. In order to extract meaningful information from high-precision data, the most significant bit (MSB) and the next few bits are often chosen. For example, quantization of an 8-bit unsigned integer number into 2-bit equals means selecting the MSB and the second-most significant bit (2nd MSB) then truncating all the other following bits. While MSBs are generally more important than 2nd MSBs for correct classification, 2nd MSBs often provide more detailed information than MSBs. Thus, in general, information from 2nd MSBs are added to that from MSBs in multi-bit neural networks. However, sometimes 2nd MSBs \textit{\textbf{alone}} may have enough importance. Figure. \ref{dog} shows an image from ImageNet dataset \cite{imagenet}. Each image is the result of quantization to (a) 8-bit, (b) 1-bit (MSB), and (c) 1-bit (2nd MSB). MSBs in the image represent the boundary between the dog and the background (Figure \ref{dog}b). On the other hand, 2nd MSBs in the image show more detailed information such as the eyes of the dog (Figure \ref{dog}c). Based on this observation, we conducted simulations to see if a network can be trained using only the 2nd MSBs of the first input feature map instead of the MSBs. Table \ref{motivation_table} shows the training results of AlexNet on ImageNet dataset. The network looking MSB is identical with the network used in XNOR-Net \cite{XNOR} and the network looking 2nd MSB is its variant where the first activation function is modified to capture 2nd MSB of the input feature map. It can be seen that the network can be trained to a reasonable level even when only the 2nd MSBs of the first feature map are used for following convolutions. This observation motivated us to construct a quantized neural network with bitwise binary activation functions. Instead of training the network with multi-bit activations, training with each bit of activations separately and then combining the results may generate similar accuracy to the conventional case. 

\begin{table}[t]
  \caption{AlexNet accuracy on ImageNet dataset.}
  \label{motivation_table}
  \centering
  \begin{tabular}{lllll}
    \toprule
    &\multicolumn{2}{c}{Train} &\multicolumn{2}{c}{Test}   \\
    \cmidrule(r){2-5}
    Look     & Top-1 & Top-5 & Top-1 & Top-5 \\
    \midrule
    MSB     & 51.01  & 75.11 & 44.97 & 69.22     \\
    2nd MSB  & 49.28  & 73.62 & 43.89 & 68.37     \\
    \bottomrule
  \end{tabular}
\end{table}

\subsection{Advantage of thresholding as binary activation function}
\label{motivation_activation}
Binary activation function has several advantages over multi-bit activation function in terms of hardware efficiency \cite{BNN,XNOR}. First of all, the activation function can be replaced by a thresholding function. From a hardware perspective, it is very easy to implement a thresholding function which serves as both activation function and quantization function. Another advantage is that several linear transformations can be combined with a thresholding function. For example, if a batch-normalization layer \cite{BN}, which is one of the linear transformations in inference phase, is followed by a thresholding layer, the two layers can be merged by adjusting the thresholds in the thresholding function. As a result, the computation cost for running  batch-normalization can be saved. Scaling of quantized weights or activations is another linear transformation required during inference. Even if binary precision is used for weight or activation, a scaling factor is often multiplied after the binary convolution. By shifting threshold values, this multiplication can also be merged with the thresholding function. In summary, most high-precision linear transformations can be merged using the thresholding function if, and only if, the binary activation function is used.

\section{Related Work}
\subsection{Extremely quantized networks}
Several methods for building a low-precision DNN have been proposed to reduce hardware complexity. Among them, BNN used only 1-bit (+1/-1) for both weights and activations \cite{BNN}. Since the sign function used in BNN is non-differentiable, ``straight-through estimator (STE)" concept was used for back-propagation \cite{STE}. BNN successfully achieved comparable accuracy with that of a full-precision network on MNIST and CIFAR-10 datasets. Rastegari et al. \cite{XNOR} proposed XNOR-Net and showed results on ImageNet dataset. XNOR-Net uses binary weights \{$+\alpha,-\alpha$\}, where $\alpha$ is the full-precision scaling factor. While BNN and XNOR-Net demonstrated reasonable accuracy results in relatively small network topologies, a considerable accuracy gap still exists between these networks and full-precision networks when the network becomes larger. To achieve higher accuracy, Zhou et. al \cite{DoReFa} proposed DoReFa-Net which used multi-bits for weights, activations and gradients. 
Recently, Cai et al. \cite{HWGQ} proposed half-wave Gaussian quantization (HWGQ). 
By using ReLU-based quantization and using only positive numbers for activations, HWGQ improved the accuracy of various deep networks on ImageNet dataset.

\subsection{Multi-bit quantized network as a bitwidth extension of BNN}
\label{abc-net}
ABC-Net proposed by Lin et al. \cite{ABC} quantized weights and activations into multiple binary codes. It was claimed in \cite{ABC} that ABC-Net is a bitwidth-extended version of BNN since ABC-Net computes multiple binary convolutions unlike the conventional multi-bit neural network that computes multi-bit convolutions. 
If successful, making bitwidth-extended version of BNN for multi-bit neural networks is very promising because it can preserve the hardware-friendly features of BNN while improving upon the classification accuracy. Despite the claims in \cite{ABC}, however, we observe that the essence of ABC-Net is re-ordering of multi-bit convolution process. In ABC-Net, all the bitwise convolution results must be added at the final stage of every convolution layer. Therefore the amount of hardware resources required to finish the convolution of $M$-bit inputs and $N$-bit weights in ABC-Net is indeed similar to that in conventional multi-bit convolution of the same inputs and weights. In addition, ABC-Net uses multi-bit activation functions that do not exploit hardware-efficient characteristics of BNN. 
Thus, we believe that a true bitwidth extension of BNN needs to have bitwise activation functions. In the next section (section \ref{BitSplit}), we propose the BitSplit-Net which is based on bitwise activation functions.

\begin{figure*}[ht]
    \centering
    \includegraphics[width=\linewidth]{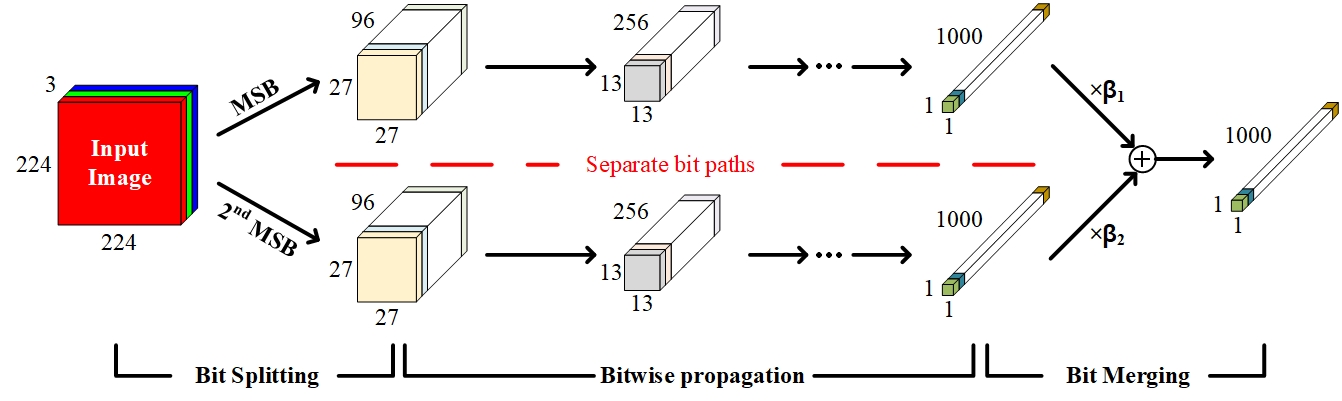}
    \caption{BitSplit version of AlexNet where each feature map is split into 2 bits: MSB and 2nd MSB.}
    \label{Network}
\end{figure*}
\section{BitSplit scheme}
\label{BitSplit}

\subsection{Overall network configuration}
\begin{figure}[t]
    \centering
    \includegraphics[width=\linewidth]{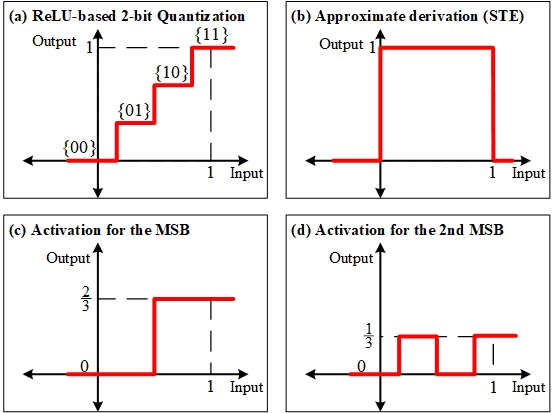}
    \caption{Activation functions used for the first layer. (a) ReLU-based 2-bit quantization function used in conventional multi-bit networks. (b) Approximate derivation of (a). (c) Activation function for the first bit in (a). (d) Activation function for the second bit in (a). The summation of (c) and (d) is identical to (a).}
    \label{Activations}
\end{figure}
In this section, we introduce the structure and characteristics of the proposed BitSplit scheme, which is based on bitwise activation function. Like existing multi-bit networks, feature map data in BitSplit-Net contain multi-bit information. However, unlike traditional multi-bit networks, each bit in the feature map propagates independently throughout the network. Note that, although each bit propagates independently, all the bits in a layer share the same weight parameters. Therefore, the size (or memory footprint) of the parameters for the network is the same as that of the original network. Figure \ref{Network} shows an example of the overall network scheme of the BitSplit version of AlexNet, where each feature map data is split into two bits. The first activation layer acts as `bit splitting layer'. The bit splitting layer splits high-precision convolution results into two paths, MSB and 2nd MSB. 
After splitting, each bit propagates separately throughout the network. At the end of the network, a `bit merging layer' merges separated bits into one feature map data. Then, the last feature map is used for the classifier. Note that, if the MSB path in the BitSplit-Net is used alone, the network becomes identical to BNN.

In BitSplit-Net, every activation function produces a 1-bit result and a simple thresholding function is used in all layers except the first activation layer. Even when 2-bit precision is used for activations, each bit goes through separate binary activation functions. In a conventional multi-bit network, convolution results from each bit are accumulated first, and then an activation function such as the one in Figure \ref{Activations}(a) is used for 2-bit activation. In contrast, convolution results from each bit are not summed together in the proposed scheme; thus preserving the efficient hardware characteristics of BNN. Based on the characteristics, we believe that BitSplit-Net extends BNN to multi-bit neural network cases better than the previous approach \cite{ABC}.

\subsection{Layers in BitSplit-Net}
In this section, we discuss the considerations required for forward and backward propagation in BitSplit-Net and provide details about each layer. For each layer, $X$ represents the input to the layer and $Y$ represents the output of the layer. Other layers that are not described below are implemented in the same way as done in conventional networks.
\paragraph{Bit splitting}
The bit splitting layer is the first layer in which BitSplit-Net starts to differ from conventional networks. In this layer, high-precision inputs are split into multiple paths by different activation functions. Figure \ref{Activations} shows how the bit splitting layer mimics a conventional ReLU-based quantization layer. Figure \ref{Activations}(a) shows a ReLU-based 2-bit quantization function when the input range is limited from 0 to 1. The 2-bit quantization produces 4 different output levels; \{00, 01, 10, 11\} indicate \{0, $\frac{1}{3}$, $\frac{2}{3}$, 1\}, respectively. In this case, the first bit has a weight of $\frac{2}{3}$ and the second bit has weight of $\frac{1}{3}$. Let us call this, the bit-weight factor, $\beta$. We designed each activation function in the bit splitting layer based on this quantization scheme and Figures \ref{Activations}(c) and (d) show two activation functions. Without loss of generality, one can design a $k$-bit splitting layer following the Algorithm \ref{algo:bitsplitting}. As described in the Algorithm \ref{algo:bitsplitting}, we constrained the activations to the range of [0,1]. In line 6, the binary code for each bit is determined and each of them is scaled by $\beta_i$ in the following line. The output of the bit splitting layer will be $\{Y_i\}_{i=1}^k = \{\beta_iY_i^B\}_{i=1}^k$ where $Y_i^B \in \{0,1\}$.

\begin{algorithm}[tb]
   \caption{Bit splitting algorithm}
   \label{algo:bitsplitting}
\begin{algorithmic}
   \STATE {\bfseries Input:} High-precision input $\mathbf{x} \in \mathbb{R}^n$, number of bits $k$
   \STATE {\bfseries Output:} $k$ binary activations $\{\mathbf{y}_i\}_{i=1}^{k}$
   \STATE Initialization: $\lambda_1 \gets 2^k -1$, $\lambda_2 \gets 0$, $\beta_i \gets 0$
   \STATE $\mathbf{x} \gets \text{ReLU1(}\mathbf{x}\text{)}\;\;\;\;\;\;\;\;\;\;\;\;\;\;\;\;\;\;\;\;\;\; \text{*equals to clamp(0,1)}$ 
   \FOR{$i=1$ {\bfseries to} $k$}
   \STATE $\lambda_2 \gets 2^{k-i}$
   \STATE $\beta_i \gets \frac{\lambda_2}{\lambda_1}$ 
   \STATE $\mathbf{y}_i \gets \text{Modulo}\Big(\text{Floor}\big(\frac{1}{\lambda_2}\text{Round(}\lambda_1 \mathbf{x}\text{)}\big)\text{,}2\Big)$
   \STATE $\mathbf{y}_i \gets \beta_i\mathbf{y}_i$
   \ENDFOR
   \STATE {\bfseries Return:} $\{\mathbf{y}_i\}_{i=1}^{k}$
\end{algorithmic}
\end{algorithm}

Since these activation functions are non-differentiable, STE concept is used for back-propagation \cite{STE}. Figure \ref{Activations}(b) is the approximate derivation of ReLU-based 2-bit activation function in (a). Similarly, gradients for each activation functions of bit splitting layer can be derived using STE. Then, gradient values from each bit are accumulated as described in Eq. \ref{eq_split_back}. Here, $\mathbf{C}$ denotes for cost function.
\begin{equation}
    \label{eq_split_back}
    \frac {\partial \mathbf{C}}{\partial \mathbf{X}} = \sum_{i=1}^k{\frac {\partial \mathbf{C}}{\partial \mathbf{Y}_i}\frac {\partial \mathbf{Y}_i}{\partial \mathbf{X}}} \;\;\myeq\;\; \sum_{i=1}^k{\frac {\partial \mathbf{C}}{\partial \mathbf{Y}_i}}\beta_i
\end{equation}

\paragraph{Bitwise convolution}
After passing through the bit splitting layer, feature maps from each bit are convoluted separately, but with same weight parameters. Let us assume \{$\mathbf{X}_i\}_{i=1}^k \in {\{0,\beta_i\}}^{w\times h\times c_{in}}$ as $k$ binary input to the convolution layer and \{$\mathbf{W}_j\}_{j=1}^n \in {\{-1,1\}}^{w\times h\times c_{in} \times c_{out}}$ as $n$-bit weight matrix. Note that, in hardware, $\{X_i^B\}_{i=1}^k \in \{0,1\}^{w\times h\times c_{in}}$ can be used instead of $\{X_i\}_{i=1}^k$, and $\beta_i$ can be combined with threshold as mentioned in section \ref{motivation_activation} without any hardware overhead. Since each bit of activation propagates independently, we do not need to use conventional multi-bit multiplication for convolutions. Instead, we utilize bitwise binary convolution for better hardware performance. As analyzed in previous works \cite{BNN,ABC,XNOR}, bitwise convolution can be substituted with simple bitwise logics such as XNOR and pop-count operations. As a result, the forward propagation of the bitwise convolution layer is given by:

\vspace{5pt}
\ Forward:
\vspace{-5pt}
\[
\mathbf{Y} = [\mathbf{Y}_1,...,\mathbf{Y}_k]\]
\begin{equation}
\mathbf{Y}_i = \sum_{l=1}^{n}{BitwiseConv(\mathbf{X}_i,\mathbf{W}_l)}
\end{equation}
The scaling factor of the weights is considered inside $BitwiseConv$ function. Note that BitSplit-Net does not require accumulation of convolution results of each activation bit while conventional multi-bit networks do. Since the way BitSplit handles multi-bit weight is same as conventional multi-bit networks, we simplified the weight matrix as $\textbf{W}$ for backward propagation.

Backward:
\vspace{5pt}
\[
    \frac {\partial \mathbf{C}}{\partial \mathbf{W}}=\sum_{i=1}^k{\frac {\partial \mathbf{C}}{\partial \mathbf{Y}_i}\frac {\partial \mathbf{Y}_i}{\partial \mathbf{W}}}=\sum_{i=1}^k{\frac{\partial \mathbf{C}}{\partial \mathbf{Y}_i}\mathbf{X}_i}
    \]
\begin{equation}\label{eq1}
    \frac {\partial \mathbf{C}}{\partial \mathbf{X}_i}=\frac {\partial \mathbf{C}}{\partial \mathbf{Y}_i}\frac {\partial \mathbf{Y}_i}{\partial \mathbf{X}_i}=\frac{\partial \mathbf{C}}{\partial \mathbf{Y}_i}\mathbf{W}
\end{equation}

Note that fully-connect layer can be considered as 1x1 convolution layer, so same approach can be applied. Similar to conventional multi-bit networks, a single weight matrix is computed with all the bits of the feature map, and gradient values from each activation bit are accumulated and used to update the weight in the training phase. 

To sum up, convolution layer in BitSplit-Net is different from that of conventional multi-bit networks in two respects. First, the computation is simpler in BitSplit case since it does not require accumulation of $\mathbf{Y}_i$s. Second, a multi-bit convolution is replaced by multiple binary convolutions. Based on these two features, BitSplit-Net can achieve better performance in various hardware platforms than the conventional low-precision neural networks; refer to section \ref{performance} for comparison results.

\paragraph{Binary activation function}
Except the first activation layer which is the bit splitting layer, all the other activation layers use the thresholding function as an activation function. While the threshold is fixed at 0.5, its output value varies depending on the bit position. The bit-weight factor, $\beta$, is again used to determine the output values of each thresholding function. For example, when $k$=2, the feature map of MSB path can have value of $0$ or $\frac{2}{3}$ and that of 2nd MSB path can have value of $0$ or $\frac{1}{3}$. This output value can be found based on the total number of bits ($k$) and which bit ($i$) the function is looking at.
For back-propagation, the STE method was used similar to the case in the bit splitting layer. 

Forward:
\[
Y_i = 
  \begin{cases}
    \beta_i & \quad \text{if}\;\; x \geq 0.5 \\
    0 & \quad \text{if}\;\; x < 0.5
  \end{cases}\\
\]

Backward:
\vspace{5pt}
\begin{equation}\label{eq2}
\frac {\partial \mathbf{C}}{\partial \mathbf{X}_i}=\frac {\partial \mathbf{C}}{\partial \mathbf{Y}_i}\frac {\partial \mathbf{Y}_i}{\partial \mathbf{X}_i}\;\;\myeq\;\; \frac {\partial \mathbf{C}}{\partial \mathbf{Y}_i} \beta_i
\end{equation}

\paragraph{Bit merging}
At the end of the network, the bit merging layer accumulates the results from each bit path to produce the last feature map for classification. The result from each bit path has a different significance; for example, the result from the MSB path may be twice as important as the result from the 2nd MSB path. However, this has been already considered by bit-weight factor, $\beta$, in the binary activation layer. Therefore, the output of the bit merging layer can be seen as the simple weighted sum of the results from each bit path. The bit-weight factors can simply be computed, or can be set to learnable parameters so that optimal values can be found during training. Please refer to section S1 of the supplementary material for a detailed method on training the bit-weight factors.

Forward:
\[
\mathbf{Y} = \sum_{i=1}^k{\beta_{i} \mathbf{X}_i^B}\\
\]

Backward:
\begin{equation}\label{eq3}
\frac {\partial \mathbf{C}}{\partial \mathbf{X}_i^B}=\frac {\partial \mathbf{C}}{\partial \mathbf{Y}}\frac {\partial \mathbf{Y}}{\partial \mathbf{X}_i^B}= \frac {\partial \mathbf{C}}{\partial \mathbf{Y}} \beta_i
\end{equation}

\subsection{Bit precision of BitSplit-Net}
In this work, we do not pursue higher than 4-bit cases because our main focus is to make it possible to use bitwise computation for multi-bit "low-precision" networks. Recently, WRPN \cite{WRPN} has shown that by doubling the number of channels, the accuracy of low-precision networks could be significantly improved and even the similar level of accuracy to the full-precision case could be achieved. Since BitSplit scheme can be applied to WRPN style, increasing the number of channels is better approach to achieve full-precision accuracy than using higher bit-resolutions. Furthermore, it gets more difficult to train BitSplit-Net as the bit precision of activation increases. Therefore, we mainly evaluated the BitSplit scheme with low-precision networks while providing few results with WRPN style to show that BitSplit scheme can benefit from WRPN approach.

In BitSplit-Net, only activations are split into binary codes. In other words, the precision of weight can be any values including binary, ternary, or even higher values such as 4-bit. We quantized weight following the methods described in previous works \cite{TWN,fixedpoint,XNOR}. Only uniform quantization was considered since it is more hardware-friendly than non-uniform quantization. Even if high precision is used for weight, the convolution operation still consists of bitwise binary convolution because the activation value is binary.

\section{Training results}
We applied the BitSplit scheme to various image classification datasets and demonstrated that BitSplit-Net is capable of achieving classification accuracy comparable to that of conventional multi-bit "low-precision" networks. We trained multiple networks including LeNet-5 \cite{lenet} for MNIST dataset, VGG-9 \cite{VGG} for CIFAR-10 dataset, and both AlexNet \cite{AlexNet} and ResNet-18 \cite{resnet} for ImageNet dataset. We trained all the networks from scratch with He initialization \cite{He}. Since a number of bitwise binary activation functions act as strong regularizer, we lowered values of the weight decay and dropout ratio. Details on hyperparameters, such as learning rate schedules and weight decay, are described in section S2 of the supplementary material. To provide a fair comparison with other works, we did not quantize the first and the last layers of the networks.

\subsection{Results on small datasets}

We first evaluated our BitSplit scheme on small datasets such as MNIST and CIFAR-10, focusing on bit-scalability. While maintaining the weight precision to binary, we observed that the classification accuracy was dependent upon the number of the activation bits. Table \ref{Small_result} shows the training results of LeNet-5 on MNIST and VGG-9 on CIFAR-10. It is clearly shown that the classification accuracy improves as the precision of activation ($k$) increases. When $k=4$, BitSplit-Net achieved an accuracy loss less than 0.5\% compared to the full-precision baseline. We also observed that the largest improvement in accuracy occurred between $k=1$ and $k=2$.
\begin{table}[t]
    \caption{Classification accuracy of BitSplit-Net for LeNet-5 on MNIST and VGG-9 on CIFAR-10. Binary weight (1-bit) was used for all cases.}
    \label{Small_result}
    \centering
    \begin{tabular}{ccccc}
        \toprule
        Network & \multicolumn{2}{c}{LeNet-5}  & \multicolumn{2}{c}{VGG-9} \\
        \cmidrule(r){2-3} \cmidrule(r){4-5} 
        & Accuracy & gap & Accuracy & gap\\
        \midrule
        Baseline & 99.32 & - & 90.36 & -\\
        k = 1 & 98.79 & \textbf{0.53} & 87.93 & \textbf{2.43}\\
        k = 2 & 99.17 & \textbf{0.15} & 89.50 & \textbf{0.86}\\
        k = 3 & 99.19 & \textbf{0.13} & 89.76 & \textbf{0.60}\\
        k = 4 & 99.22 & \textbf{0.10} & 89.99 & \textbf{0.37}\\
        \bottomrule
    \end{tabular}
\end{table}

\subsection{Results on ImageNet dataset}
We also evaluated the BitSplit version of AlexNet and ResNet-18 on an ImageNet dataset which is a larger dataset than MNIST or CIFAR-10. 
We trained the BitSplit version of AlexNet and ResNet-18 with different bit configurations and compare the classification accuracy with other low-precision networks. Table \ref{AlexNet_result} and Table \ref{ResNet18_result} show the Top-1 and Top-5 accuracy and accuracy gap to the full-precision baseline for AlexNet and ResNet-18, repectively. We compared BitSplit-Net with other quantized networks where uniform quantizaion is used and ABC-Net which used nonlinear quantization. BitSplit versions of AlexNet and ResNet-18 were successfully trained to have similar accuracy level ($<\pm 0.5\%$) compared to other existing networks including ABC-Net. In addtion, increasing the precision of activation or weight resulted in more improvement on classification accuracy. ``BitSplit 2x" in the last row of each table shows the training result when WRPN style was applied. The classification accuracy could be improved to the similar level with baseline when WRPN method was applied.

\begin{table}[t]
  \caption{Classification accuracy of AlexNet in various schemes.}
  \label{AlexNet_result}
  \centering
  \begin{tabular}{ccccccc}
    \toprule
    &\multicolumn{2}{c}{Precision} &\multicolumn{4}{c}{Accuracy}   \\
    \cmidrule(r){2-3} \cmidrule(r){4-7}
    Network     & A & W & Top-1 & Top-5 & Gap-1 & Gap-5\\
    \midrule
    Baseline & FP & FP & 57.1 & 80.2 & - & -\\
    BNN     & 1  & 1 & 41.8 & 67.1 &  15.3 &  13.1  \\
    XNOR  & 1  & 1 & 44.2 & 69.2 & 12.9 &  11.0   \\
    \multirow{2}{*}{DoReFa} 
    & 2 & 1 & 47.7 & - & 9.4  & -\\
    & 3 & 1 & 48.4 & - & 8.7 & -\\
    QNN & 2 & 1 & 51.0 & 73.7 & 6.1 & 6.5 \\
    \multirow{2}{*}{HWGQ} & 2 & 1 & 50.5 & 74.6 & 6.6 & 5.6\\
    & 3 & 1 & 51.9 & 75.7 & 5.2 & 4.5 \\
    \midrule
    \multirow{5}{*}{\textbf{BitSplit}} & 2 & 1 & 51.0 & 74.5 & 6.1 & 5.7\\
    & 3 & 1 & 51.8 & 75.3 & 5.3 & 4.9\\
    & 2 & 2 & 52.6 & 76.3 & 4.5 & 3.9\\
    & 3 & 2 & 54.0 & 77.3 & 3.1 & 2.9\\
    & 2 & 4 & 54.8 & 77.8 & 2.3 & 2.4\\
    \midrule
    \textbf{BitSplit} &\multirow{2}{*}{2} & \multirow{2}{*}{4} & \multirow{2}{*}{56.3} & \multirow{2}{*}{78.8} & \multirow{2}{*}{0.8} &\multirow{2}{*}{1.4}\\
    \textbf{2x} & & & & & & \\
    \bottomrule
  \end{tabular}
\end{table}
\vspace{-2mm}
\begin{table}[t]
  \caption{Classification accuracy of ResNet-18 in various schemes.}
  \label{ResNet18_result}
  \centering
  \begin{tabular}{ccccccc}
    \toprule
    &\multicolumn{2}{c}{Precision} &\multicolumn{4}{c}{Accuracy}   \\
    \cmidrule(r){2-3} \cmidrule(r){4-7}
    Network     & A & W & Top-1 & Top-5 & Gap-1 & Gap-5 \\
    \midrule
    Baseline & FP & FP & 69.3 & 89.2 & - & -\\
    BNN     & 1  & 1 & 42.2 & 67.1 & 27.1 & 22.1   \\
    XNOR  & 1  & 1 & 51.2 & 73.2 & 18.1 & 16.0     \\
    HWGQ & 2 & 1 & 56.1 & 79.7 & 13.2 & 9.5\\
    \multirow{2}{*}{ABC-Net} & 1 & 1 & 42.7 & 67.6 & 26.6 & 21.6\\
    & 3 & 3 & 61.0 & 83.2 & 8.3 & 6.0 \\
    \midrule
    \multirow{6}{*}{\textbf{BitSplit}} & 2 & 1 & 56.8 & 79.7 & 12.5 & 9.5\\
    & 2 & 2 & 58.4 & 80.6 & 10.9 & 8.6\\
    & 2 & 3 & 59.0 & 81.3 & 10.3 & 7.9\\
    & 3 & 2 & 59.2 & 81.4 & 10.1 & 7.8\\
    & 2 & 4 & 60.6 & 82.5 & 8.7 & 6.7\\
    & 3 & 3 & 61.2 & 82.8 & 8.1 & 6.4\\
 \midrule
   \textbf{BitSplit} &\multirow{2}{*}{2} & \multirow{2}{*}{4} & \multirow{2}{*}{67.9} & \multirow{2}{*}{88.0} & \multirow{2}{*}{1.4} &\multirow{2}{*}{1.2}\\
    \textbf{2x} & & & & & & \\
    \bottomrule
  \end{tabular}
\end{table}

\section{Hardware benefit of BitSplit-Net}
\label{performance}
\subsection{Bitwise dot-product with \{0,1\} activation and \{-1,1\} weight}
\label{XNOR-AND}
In this section, we first would like to explain how we handle bitwise binary convolution with \{0,1\} activations and \{-1,1\} weights in hardware level. Similar to the fully-connected layer, the main operation of convolution is also dot-product. In previous works \cite{BNN,ABC,XNOR}, it was proposed that binary dot-product of activation $\mathbf{X} \in {\{-1,+1\}}^{1\times N}$ and weight $\mathbf{W} \in {\{-1,+1\}}^{N\times 1}$ can be computed using XNOR and pop-count operations. Since 1-bit digital numbers can only have 0 or 1 value, $\mathbf{X}$ and $\mathbf{W}$ have to be encoded to $\mathbf{X}^D \in {\{0,1\}}^{1\times N}$ and $\mathbf{W}^D \in {\{0,1\}}^{N\times 1}$. Then the binary dot-product can be computed as follows:
\vspace{-3mm}

\begin{equation}
    \label{XNOR-popcount}
    \vspace{-2mm}
    \begin{aligned}
        &X\cdot W \\
        &= N -2\left(\sum_{i=1}^{N/64}{POPCNT(XNOR(X_i^D,W_i^D))}\right)
    \end{aligned}
\end{equation}
Here, the number of inputs for pop-count operation is fixed to 64. However, when ReLU-based activation function is used for activation quantization \cite{HWGQ}, activation values are 0 or 1, not -1 or +1, therefore we need to modify the Eq. \ref{XNOR-popcount} as follows.

\begin{equation}
    \label{AND-popcount}
    \begin{aligned}
    &X\cdot W\\
    &=2\left(\sum_{i=1}^{N/64}{POPCNT(AND(X_i^D,W_i^D))}\right)\\
    &\;\;\;\;\;-\sum_{i=1}^{N/64}{POPCNT(X_i^D)}
    \end{aligned}
\end{equation}
Similar to Eq. \ref{XNOR-popcount}, the dot-product of \{0,1\} activations and \{-1,+1\} weights can be computed using AND logic and pop-count operations (Eq. \ref{AND-popcount}). To consider the number of `0's, another pop-count is needed. However, this additional pop-count can be performed only once and then reused many times during the dot-product. Therefore, the computational cost of XNOR-based dot-product and that of AND-based dot-product are almost same.

\subsection{BitSplit CUDA kernel for GPU acceleration}

\begin{figure}[t]
    \centering
    \includegraphics[width=\linewidth]{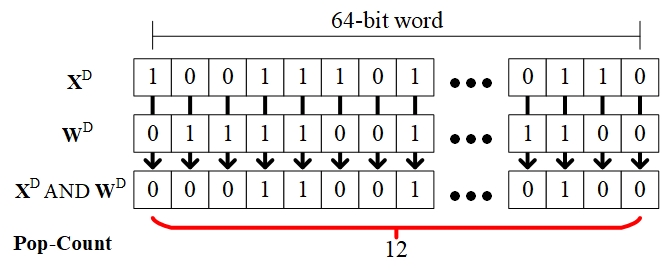}
    \caption{Dot-product using AND and pop-count operations in GPU architecture. }
    \label{AND-POPCNT}
\end{figure}

Graphics Processing Unit (GPU) is one of the most powerful and popular hardware platforms for DNN training and inference. GPU is known to be efficient for parallel data processing such as matrix multiplication, therefore is very suitable for DNN acceleration. To evaluate the benefit of BitSplit scheme on the state-of-art GPU architecture, we implemented a custom CUDA kernel for BitSplit layers. We utilized the Espresso framework \cite{Espresso} which is a CUDA kernel for BNN and modified it to cover multi-bit BitSplit-Net. Since BitSplit-Net can exploit bitwise convolution for multi-bit cases, we can leverage the key features of Espresso including bit-packing and efficient memory management. Figure \ref{AND-POPCNT} shows how binary dot-product is performed in GPU architecture. 64 binary values are packed in a single 64-bit word and used for computations. The first stage is the bitwise AND operation using two 64-bit words. The following stage is counting the number of `1's in 64-bit results. Each of both stages can be supported by a single instruction in the state-of-the-art GPU architectures.

We evaluate the BitSplit kernel in two stages. First we compared the time consumed to finish the multiplication of a matrix (8192x8192) and a vector (8192x1) which is a key computation in neural network. Then, we also compared the total time elapsed to compute forward propagation of MLP network on MNIST benchmark. Conventional multi-bit networks always take fixed amount of time to finish those benchmarks regardless of the bit precision since GPUs do not support low-precision multiplication. The low-precision weights and activations are usually encoded to higher precision such as single precision (32-bit floating point, FP32) and then computed using floating point multipliers. Therefore there is no difference in speed between networks with different bit precisions. In most of state-of-the-art GPU architectures, 16-bit floating point (FP16) multiplication or integer 8-bit (INT8) multiplication are also supported. However, many of deep learning frameworks do not fully support FP16 and INT8 operations yet. Therefore most of low-precision networks use single precision to store their low-precision data. In this context, we used single precision for baseline experiment.

The speed-up of BitSplit kernel on matrix multiplication stems from two different technical aspects: (1) memory reduction, (2) computational gain. We first introduce the details on each aspect and then present the experimental data using MNIST benchmark.

\paragraph{Memory reduction} Since BitSplit-Net can exploit bitwise computation for multi-bit convolution (or fully-connected) layers, the bit-packing technique introduced in Espresso~\cite{Espresso} can be utilized. By using bit-packing, we can reduce the memory requirement (i.e. 32 times memory reduction when binary precision is used). Assuming that we have 32 binary data, conventional kernel assigns each binary data into 32-bit floating point format, which requires 1,024 bits to store the data. In bit-packing method, 32 binary data are packed and stored in a 32-bit storage without any wasted memory usage. This technique reduces the memory access time during operations as well as the memory requirement. Since 32 binary data can be accessed at the same time, we can achieve 32x faster memory access than conventional kernel with 32-bit floating point format.

\paragraph{Computational gain} Bitwise computation also provides computational gain in addition to memory reduction. As described in section 6.1, bitwise dot-product can be computed with bitwise logic operations and pop-count operations. Even though high precision Multiply-and-Accumulate (MAC) operation is faster than XNOR (or AND) and pop-count operations, the throughput is much higher with XNOR and pop-count operations since 32 data are computed at the same time. 
 
\subsection{Speed-up analysis of BitSplit kernel}

\begin{figure}[t]
    \centering
    \includegraphics[width=\linewidth]{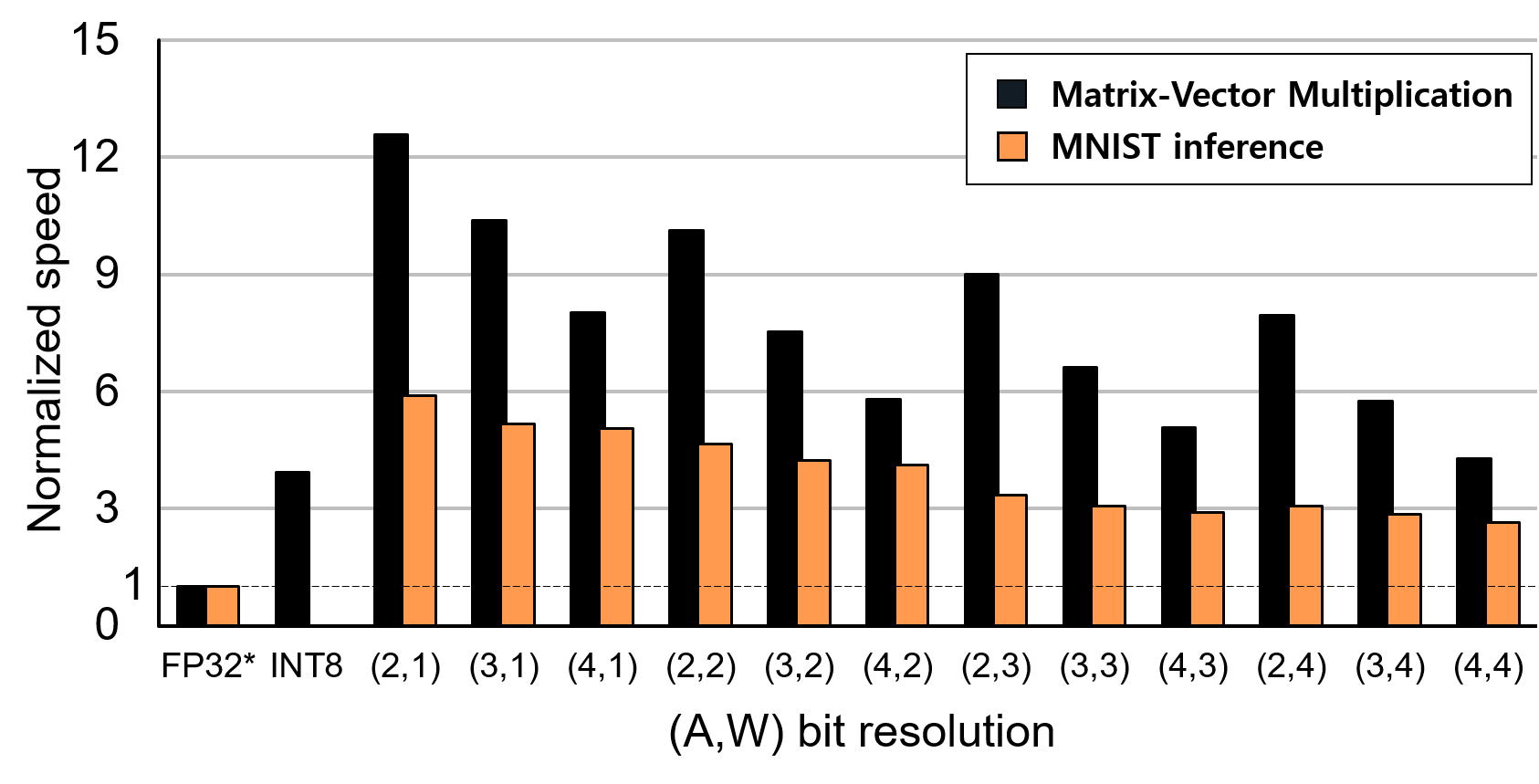}
    \caption{Normalized speed of matrix-vector multiplication (left) and forward propagation of MLP on MNIST benchmark (right).}
    \label{speed_comparison}
\end{figure}

Based on these technical advantages, BitSplit kernel can achieve higher computing speed as shown in Figure \ref{speed_comparison}. The left bar in each bin shows the normalized speed of matrix-vector multiplication with different bit configurations. Each result was normalized to the speed of FP32 case which is our baseline. For fair comparison, same level of optimization techniques were used for all the results except the FP32 case where cuBLAS library was used. In any configuration, as shown in Figure \ref{speed_comparison}, BitSplit kernel is much faster than FP32 case. It is also clearly shown that the speed-up of the kernel is reduced as the bit resolution increases. BitSplit kernel with 2-bit activation and 1-bit weight was 12.5x faster than the baseline kernel. BitSplit kernel with 4-bit activation and 4-bit weight was still 4.26x faster than the baseline kernel and also faster than INT8 kernel.

\begin{figure}[t]
    \centering
    \includegraphics[width=\linewidth]{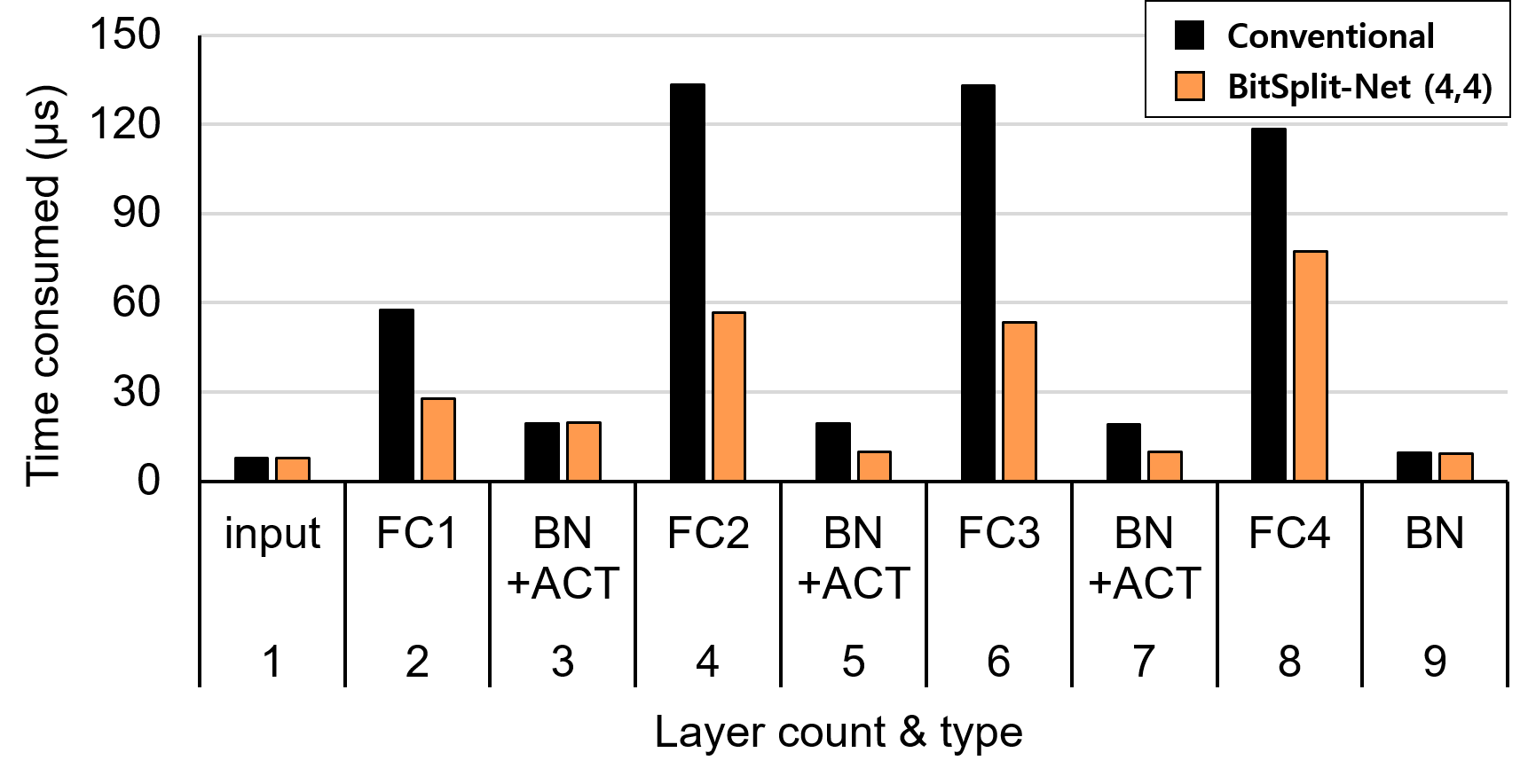}
    \caption{Time consumed for each layer in conventional scheme (left) and BitSplit-Net with 4-bit/4-bit configuration (right).}
    \label{layer_breakdown}
\end{figure}

The speed-up of BitSplit version of MLP on MNIST benchmark is also shown in Figure \ref{speed_comparison}. The network used for the benchmark is same as the one described in \cite{BNN}. The speed of inference using BitSplit kernel was much higher than that of FP32 case in this case, too. Figure \ref{layer_breakdown} shows the time consumed for each layer in the MLP network. In conventional network, computing for fully-connected layers dominates the computation time (80\%). As mentioned earlier, BitSplit kernel can boost computation of these layers substantially, therefore inference time can also be significantly reduced. Moreover, BitSplit-Net can benefit from bitwise propagation throughout the network. In conventional multi-bit networks, high-precision computations such as batch-normalization have to be done first and then multi-bit activation layer can be computed. In BitSplit-Net, however, these layers can be merged so that large amount of time is saved. As a result, BitSplit-Net can achieve 5.89x speed-up with 2-bit activation and 1-bit weight, and 2.65x speed-up with 4-bit/4-bit compared to the conventional case using FP32 format. The reason why speed-up numbers for inference time is smaller than that for matrix-vector multiplication is that the size of the benchmark network is too small to experience the benefit of BitSplit kernel. Since state-of-the-art networks for real-world image classification tasks usually require much larger matrix-vector multiplication, we believe that speed-up of BitSplit-Net will become higher with larger networks.

\section{Conclusion}
We propose a quantized neural network called BitSplit-Net, where each bit propagates over the separate network path. We believe that BitSplit-Net is a better form of bitwidth extension of BNN than the previous works because a multi-bit activation is split to multiple binary activations in the proposed scheme. We show that BitSplit scheme has a good bit-scalability for the quantized neural networks with low precision ($\leq$ 4-bit) as the increased number of the activation bits leads to improvement of classification accuracy. We further show that BitSplit version of a network can achieve similar accuracy to that of conventional multi-bit networks for several datasets. We also implement the proposed BitSplit scheme on state-of-the-art GPU platforms. We implement custom CUDA kernel for BitSplit-Net in which several advantages of BitSplit scheme can be utilized. We evaluated the BitSplit kernel on matrix-vector multiplication benchmark and forward propagation of MLP on MNIST benchmark. Experimental results show that BitSplit-Net can achieve 4.26x$\sim$12.5x speed-up on matrix-vector multiplication compared to cuBLAS library. Furthermore, 2.65$\sim$5.89x speed-up on MNIST inference was achieved in comparison to conventional multi-bit scheme using FP32 format. Since BitSplit-Net inherits the hardware-friendly and light-weight characteristics of BNN, it also has high potential to be used in the resource-limited applications. Overall results show that the proposed BitSplit scheme can achieve better hardware efficiency while maintaining comparable accuracy level to that of previous multi-bit networks.

\section*{Acknowledgements}
This work was in part supported by the Technology Innovation Program (10067764) funded by the Ministry of Trade, Industry \& Energy(MOTIE, Korea) and Samsung Electronics Co., Ltd. This work was also in part supported by MSIT(Ministry of Science and ICT), Korea, under the ICT Consilience Creative program(IITP-2019-2011-1-00783) supervised by the IITP(Institute for Information \& communications Technology Planning \& Evaluation).

\newpage
\onecolumn
{\Huge\center Supplementary material}

\vspace{5mm}
\section*{S1 Training bit-weight factor $\beta$}
In BitSplit-Net, each bit path has different weight factor, $\beta$. This bit-weight factor affects the network in several ways. For example, when all the results from each bit path are summed, they are weighted-summed by the factor of $\beta_i$s. $\beta$ can be simply set to fixed value as described in Algorithm 1, or can be set to a trainable parameter. During the training phase, $\beta$s can be can be updated as follows:
\begin{equation*}
    \label{eq_beta}
    \frac {\partial \mathbf{C}}{\partial \beta_i}=\frac {\partial \mathbf{C}}{\partial \mathbf{Y}}\frac {\partial \mathbf{Y}}{\partial \beta_i}= \frac {\partial \mathbf{C}}{\partial \mathbf{Y}} \mathbf{X}_i
\end{equation*}
Since small change of $\beta$s can affect the network substantially, we used 10 times smaller learning rate for $\beta$s than other parameters. Furthermore, $\beta$s in other layers can also be trained in similar way to have different values in each layer. Again, this does not imply any hardware overhead since $\beta$s can be merged with thresholds of the following activation layer. In our experiments, we found fixed $\beta$s work reasonably, therefore used fixed $\beta$s to make optimization problem simple.

\section*{S2 Parameter settings for the experiments in section 5}
Here we give detail information on hyperparameter settings used for the experiments in section 5. Table S1 shows summary of the parameter settings. Since a number of bitwise binary activation functions act as strong regularizer, we lowered the weight decay and dropout ratio. Learning rate decay epoch indicates at which epoch learning rate was decayed. For exampe, in case of VGG-9, the initial learning rate of 0.005 was used until 80th epoch and then multiplied by 0.1. After 20 more epochs, the learning rate was again multiplied by 0.1 therefore it becomes 0.00005. The two values of learning rate decay in AlexNet case were used alternatively at every decay epoch.
\begin{table*}[!h]
  \renewcommand\thetable{S1}
  \caption{Hyper-parameter settings for different datasets.}
  \label{parameter_setting}
  \centering
  \begin{tabular}{ccccc}
    \toprule
    &MNIST & CIFAR-10 & \multicolumn{2}{c}{ImageNet}   \\
    \midrule
    Network     & LeNet-5 & VGG-9 & AlexNet & ResNet-18B \\
    \midrule
    Weight decay & $10^{-5}$ & $10^{-5}$ & $10^{-5}$ & $10^{-7}$ \\
    Dropout ratio & -  & 0.1 & 0.1 & -  \\
    Batch size  & 100  & 256 & 256 & 256  \\
    Initial learning rate & 0.1 & 0.005 & 0.0005 & 0.005\\
    Learning rate decay epoch & 15, 30, 45 & 80, 100 & 20, 30, 40, 50 & 30, 60, 80\\
    Learning rate decay & 0.5 & 0.1 & 0.5, 0.2 & 0.1 \\
    Optimizer & SGD & ADAM & ADAM & ADAM\\
    Momentum & 0.9 & - & - & - \\
    Total epoch & 50 & 120 & 60 & 90\\
    \bottomrule
  \end{tabular}
\end{table*}

\end{document}